\documentclass[fleqn,10pt]{wlscirep}%, twocolumn

\usepackage{graphicx}
\usepackage{caption}
\usepackage{subcaption}
\usepackage{booktabs}
\usepackage{amsmath}
\usepackage[normalem]{ulem} % either use this (simple) or
\usepackage{soul} % use this (many fancier options)

\title{Towards radiologist-level cancer risk assessment\\ in CT lung screening using deep learning}

\author[1,*,+]{Stojan Trajanovski}
\author[1,*,+]{Dimitrios Mavroeidis}
\author[2,*]{Christine Leon Swisher}
\author[1]{\\Binyam Gebrekidan Gebre}
\author[3,1]{Bastiaan S. Veeling}
\author[4]{Rafael Wiemker}
\author[4]{Tobias Klinder}
\author[5]{\\Amir Tahmasebi}
\author[6]{Shawn M. Regis}
\author[6]{Christoph Wald}
\author[6]{Brady J. McKee}
\author[6]{\\Sebastian Flacke}
\author[7]{Heber MacMahon}
\author[5]{Homer Pien}

\affil[*]{The first three authors have contributed equally.}
\affil[+]{Corresponding authors: \{stojan.trajanovski, dimitrios.mavroeidis\}@philips.com.}
\affil[1]{S.T., D.M., and B.G.G. are with Data Science department, Philips Research, 5656 AE Eindhoven, The Netherlands.}
\affil[2]{C.L.S. is now with Human Longevity, Inc., San Diego, CA 92121, USA. The research was done while C.L.S. was with Philips Research North America, Cambridge, MA 02141, USA.}
\affil[3]{B.S.V. is with Machine Learning lab, University of Amsterdam, 1090 GH Amsterdam and with Philips Research, 5656 AE Eindhoven, The Netherlands.}
\affil[4]{R.W. and T.K. are with Digital Imaging department, Philips Research, 22335 Hamburg, Germany.}
\affil[5]{A.T. and H.P. are with Philips Research North America, Cambridge, MA 02141, USA.}
\affil[6]{S.M.R., C.W., B.J.M., and S.F. are with Lahey Hospital \& Medical Center, Burlington, MA 01805, USA.}
\affil[7]{H.M. is with is with the Department of Radiology, University of Chicago, Chicago, IL 60637, USA.}

\begin{abstract}
\end{abstract}
\begin{document}

\flushbottom
\maketitle

\vspace{-2.6cm}

\section*{Key Points}
\paragraph{Question:} How does a deep learning model for Lung Cancer Screening perform as compared to radiologists, how well do deep learning algorithms generalize across multiple datasets and how should we train these algorithms?

\paragraph{Finding:} A deep learning model demonstrates stable performance across three low-dose lung cancer screening datasets (NLST, LHMC, and Kaggle competition data), better performance than the widely used PanCan model, improved performance compared to the state-of-the-art represented by the winners of the Kaggle Data Science Bowl challenge on lung cancer screening, and comparable performance to a panel of six radiologists. Our findings also demonstrate the importance of a good nodule detector and confirms that the training for nodule detection and malignancy score prediction can be two separate processes.

\paragraph{Meaning:} The results suggest that a deep learning algorithm may be helpful to radiologists in their practice as a decision support tool or second opinion, but this will require further validation in a clinical setting.

\section*{Abstract}
\paragraph{Importance:} Lung cancer is the leading cause of cancer mortality in the US, responsible for more deaths than breast, prostate, colon and pancreas cancer combined and it has been recently demonstrated that low-dose computed tomography (CT) screening of the chest can significantly reduce this death rate.

\paragraph{Objective:} To compare the performance of a deep learning model to state-of-the-art automated algorithms and radiologists as well as assessing the robustness of the algorithm in heterogeneous datasets.

\paragraph{Design, Setting, and Participants:} Three low-dose CT lung cancer screening datasets from heterogeneous sources were used, including National Lung Screening Trial (NLST, n=3410), Lahey Hospital and Medical Center (LHMC, n=3174) data, Kaggle competition data (from both stages, n=1595+505) and the University of Chicago data (UCM, a subset of NLST, annotated by radiologists, n=197). Relevant works on automated methods for Lung Cancer malignancy estimation have used significantly less data in size and diversity. At the first stage, our framework employs a nodule detector; while in the second stage, we use both the image area around the nodules and nodule features as inputs to a neural network that estimates the malignancy risk for the entire CT scan. We trained our algorithm on a part of the NLST dataset, and validated it on the other datasets, ensuring there is no patient overlap between the train and validation sets.

\paragraph{Exposures:} Two stage deep learning algorithms for lung cancer malignancy assessment.

\paragraph{Main outcome measures:} A binary classifier for malignancy assessment of a low-dose CT lung cancer screening exam at a patient level, evaluated by Area Under receiver operating characteristic Curve (AUC) in the three datasets.

\paragraph{Results, Conclusions, and Relevance:} The proposed deep learning model: (a) generalizes well across all three data sets, achieving AUC between 86\% to 94\%; (b) has better performance than the widely accepted PanCan Risk Model, achieving 11\% better AUC score; (c) has improved performance compared to the state-of-the-art represented by the winners of the Kaggle Data Science Bowl 2017 competition on lung cancer screening; (d) has comparable performance to radiologists in estimating cancer risk at a patient level. Moreover, the results demonstrate that nodule detection and malignancy assessment can be two independent processes.

\section*{Introduction}
Lung cancer is the leading cause of cancer mortality in the US, responsible for more deaths than breast, prostate, colon and pancreas cancer combined~\cite{Siegel_etal_2014}. Average five year survival for lung cancer is approximately 18.1\% (see e.g.~\cite{NAACCReview}), much lower than other cancer types due to the fact that symptoms of this disease usually only become apparent when the cancer is already at an advanced stage. However, early stage lung cancer (stage I) has a five-year survival of 60-75\%. Recently, the National Lung Screening Trial (NLST) demonstrated that lung cancer mortality can be reduced by at least 20\% using a screening program of high-risk populations with low-dose computed tomography (CT) of the chest annually~\cite{NLST_2011}. A greater mortality reduction has been recently reported in the European Nelson trial~\cite{IASLC18_abstracts}. More complete implementation of such lung cancer screening (LCS) programs at a national level could result in a large volume of CT lung screening (CTLS) scans that need to be assessed by radiologists since the insurance mandated high-risk criteria are met by millions of Americans~\cite{deKoning_et_al}. This highlights the potential utility of image analysis tools that can help radiologists to assess malignancy risk associated with a CTLS scan and make recommendations for further work-up, e.g. pulmonologic, oncologic or surgical evaluation.

Initial research papers on automated nodule detection were published almost 3 decades ago~\cite{Preteux1991, ShihChung_etal}, with most of the efforts focusing on chest-radiography and CT scans. More recent works~\cite{Messay_et_al2010,Camarlinghi_et_al_2012} have employed machine learning methods such as combinations of several classifiers~\cite{SuarezCuenca_etal_2011,Murphy_et_al_2009}. In a recent data challenge (Lung Nodule Analysis \emph{LUNA16}~\cite{LUNA16}), the best performing algorithms have been almost exclusively based on deep learning. Several neural network architectures rely on 2D and 3D convolutional networks to detect nodules (see e.g.,~\cite{Dou_et_al_2017, Shen_et_al_2016, Setio_et_al_2016,Li_et_al_2016,Sun_et_al_2016,Teramoto_et_al_2016,Shin_et_al_2016,Anirudh_et_al_2016}). On the topic of nodule malignancy estimation, several recent works rely (at least partially) on deep learning, e.g.~\cite{Litjens_2017,Ciompi_etal_2015,vanGinneken_etal_2015,Hua_et_al_2015,Setio_et_al_2016,Cheng_Dong_2016,Shen_etal_2015,Chen_et_al_2017,vanRiel_et_al2017,Ciompi_etal_2017}. The majority of previous studies evaluate the malignancy at the nodule level, rather the patient (full volume) level which is considered in this paper. This is beneficial for two reasons, firstly, it allows us to train a model with less information, using only patient outcomes and secondly, it allows us to evaluate the consequences of model performance at patient level. McWilliams \emph{et al.}~\cite{McWilliams_Vancouver_2013} have proposed the \emph{PanCan risk model} that uses 9 image and patient features in order to estimate the cancer malignancy risk of a nodule. Moreover, a recent study~\cite{vanRiel_et_al_2017} has shown that the PanCan Risk Model performs better than Lung-RADS\textsuperscript{TM}~\cite{Lung-RADS} and the National Comprehensive Cancer Network (NCCN) Lung Cancer Screening Guidelines version 1.2016~\cite{NCCN_guidelines}, hence highlighting it as a useful benchmark model for our experimental validation.

Recently, a data science competition (Kaggle, Data Science Bowl 2017~\cite{Kaggle_dsb2017}) was organized on the topic of lung cancer screening that attracted a significant attention from the research community. The authors of the winning entry~\cite{grt123team_kaggle2017} proposed a 3D convolutional network for nodule detection, using LUNA16 dataset and additional manual nodule annotations of the Kaggle dataset to train their nodule detector. Subsequently, five detected nodules were used as inputs for the malignancy risk assessment network.

In this paper, we propose a two-stage Machine Learning framework that estimates the cancer risk associated with a given CTLS scan. The first stage employs a detector~\cite{Bergtholdt_etal_Hamburg_detector_2016,grt123team_kaggle2017} that identifies nodules contained within a scan. The second stage employs a neural network inspired by the ResNet architecture~\cite{He_etal_2016}, regularized using dropout~\cite{JMLR:v15:srivastava14a} that performs the cancer risk assessment of the whole CTLS scan. Our framework is evaluated against three criteria (i) Robustness: we show that our framework achieves consistent performance across different low-dose CT datasets, (ii) Performance against state-of-the-art (SOTA): we show that our framework has improved performance over state-of-the-art models, and (iii) Performance compared to radiologists: we show that our model has comparable performance to a panel of six radiologists.

\section*{Methods}
\subsection*{Data Sets}

In order to evaluate the performance of our framework, we use CTLS datasets from several sources: NLST~\cite{NLST_2011}, LHMC, Kaggle~\cite{Kaggle_dsb2017} (from both competition stages) and University of Chicago (UCM) data (NLST subset with radiologist annotations). The main data characteristics are summarized in Table~\ref{table:Data_used}. The CTLS datasets we have used in our analysis come from heterogeneous sources (different hospitals, image quality, reconstruction filters, etc.) and allowed us to validate the generalization capacity of our framework. We should note that for the NLST dataset we have used the diagnosed cancer CTLS scans, but only a subset of the benign cases. It should be noted that when validating our model we ensured that there was no patient overlap between the train and validation sets, for example by removing the (cancer and non-cancer) patients that are contained in the UCM dataset. 

\begin{table*}[h!tb]
\centering
\caption{Data used in our analysis.}
\label{table:Data_used}
\resizebox{\textwidth}{!}{\begin{tabular}{l|rrr|ccc}
         & \multicolumn{3}{|c|}{Number of volumes} &     \multicolumn{3}{c}{Metadata}                                                              \\
\multicolumn{1}{c|}{Dataset}         & \multicolumn{1}{c}{Total} & \multicolumn{1}{c}{positive} & \multicolumn{1}{c|}{train or valid.}  & \multicolumn{1}{c}{nodule annotations} & \multicolumn{1}{c}{Lung-RADS\textsuperscript{TM} classification} & \multicolumn{1}{c}{radiologists' scores} \\
\toprule\toprule
NLST & 3410                     & 680                     & train  & yes & no & no                                   \\
LHMC      & 3174                   & 56                   & valid        & yes & yes & no              \\
UCM      & 197                   & 64                   & valid       & yes & no & yes (for 99 volumes)                \\
Kaggle (stage 1) train      & 1397                   & 362                   & train model~\cite{grt123team_kaggle2017}     & no & no  & no                  \\
Kaggle (stage 1) valid      & 198                   & 57                   & valid     & no & no  & no                  \\
Kaggle (stage 2)      & 505                   & 153                   & valid  & no & no  & no\\  
\toprule                                     
\end{tabular}}
\end{table*}

% add Table 1 here

We trained our model on the NLST data~\cite{NLST_2011} (3410 volumes, containing 680 hundred diagnosed cancer cases) since this dataset has the largest number of cancer cases. The NLST-trained model was subsequently validated with the other datasets. When validating model performance for the UCM dataset, we always excluded from the NLST training data the cancer and non-cancer patients that were included in the UCM study. The lung cancer screening dataset provided by LHMC contains 3174 CTLS patient scans (with 56 cancer cases), along with a nodule lexicon table that contains detailed information about the identified nodules (such as size, location, etc.). There is only a small number of cancer cases in the LHMC dataset, but the detailed nodule information allows us to compare our framework with other models from the literature that rely on such nodule-level information~\cite{McWilliams_Vancouver_2013,Lung-RADS}. Furthermore, UCM has provided additional annotations for 197 volumes of the NLST data (that contain 64 cancer cases), that allow us to compare our model with radiologists' assessment as well as the \emph{PanCan risk model}. Finally, we use the data from both stages of a recent lung cancer competition (National Data Science Bowl 2017) organized and hosted by Kaggle~\cite{Kaggle_dsb2017}. We should note that the origin of the Kaggle dataset was not disclosed by the competition organizers. Therefore, we cannot exclude the possibility that our models that have been trained on the NLST may have an overlap with the Kaggle data. This should be taken into account when interpreting the Kaggle results only. In the first stage of the competition, 1397 CTLS volumes were provided for training data (with 362 diagnosed cancer cases) and for validation 198 CTLS volumes (with 57 diagnosed cancer cases), while in the second stage 505 volumes were provided (with 153 cancer cases). In all our datasets, cancer cases were confirmed with diagnostic tests (like biopsy), so it is almost certain that the labeling is unambiguous, however, for the non-cancer cases there is a possibility that a patient left the study and developed cancer later on.

\subsection*{Machine Learning Framework for Cancer Risk Assessment}

\begin{figure*}[h!tb]
    \centering
%    \begin{subfigure}[b]{0.32\textwidth}
%        \centering
%        \includegraphics[width=\textwidth]{./images/ucm_roc_ourmodel_radiologists_allsamples.png}
%        \caption{}\label{fig:UCM_ourmodel_vs_radiologists_ROC_allsamplesourmodel}
%    \end{subfigure}
    \begin{subfigure}[b]{0.85\textwidth}
        \centering
\caption{}\label{fig:Our_algorithm_pipeline1}
        \includegraphics[width=1.058\textwidth,trim=90mm 70mm 60mm 65mm, clip=true,scale=2.0]{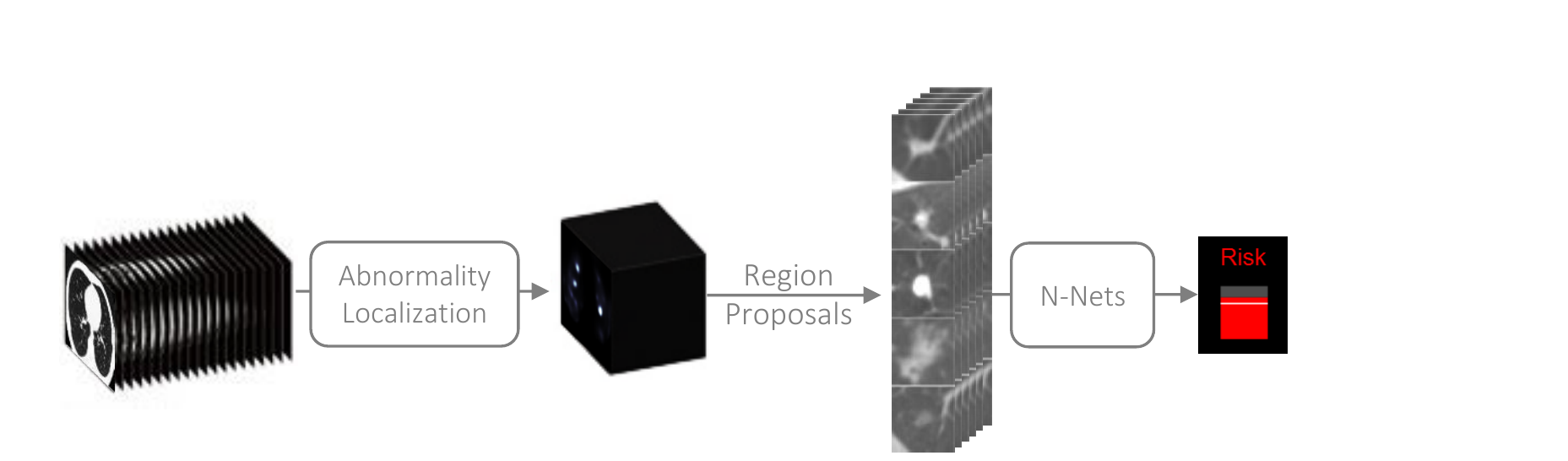}
    \end{subfigure}

		\begin{subfigure}[b]{0.85\textwidth}
        \centering
\caption{}\label{fig:DNN_nodule2cancer_prediction}
        \includegraphics[width=1.058\textwidth,trim=100mm 60mm 70mm 55mm, clip=true,scale=2.0]{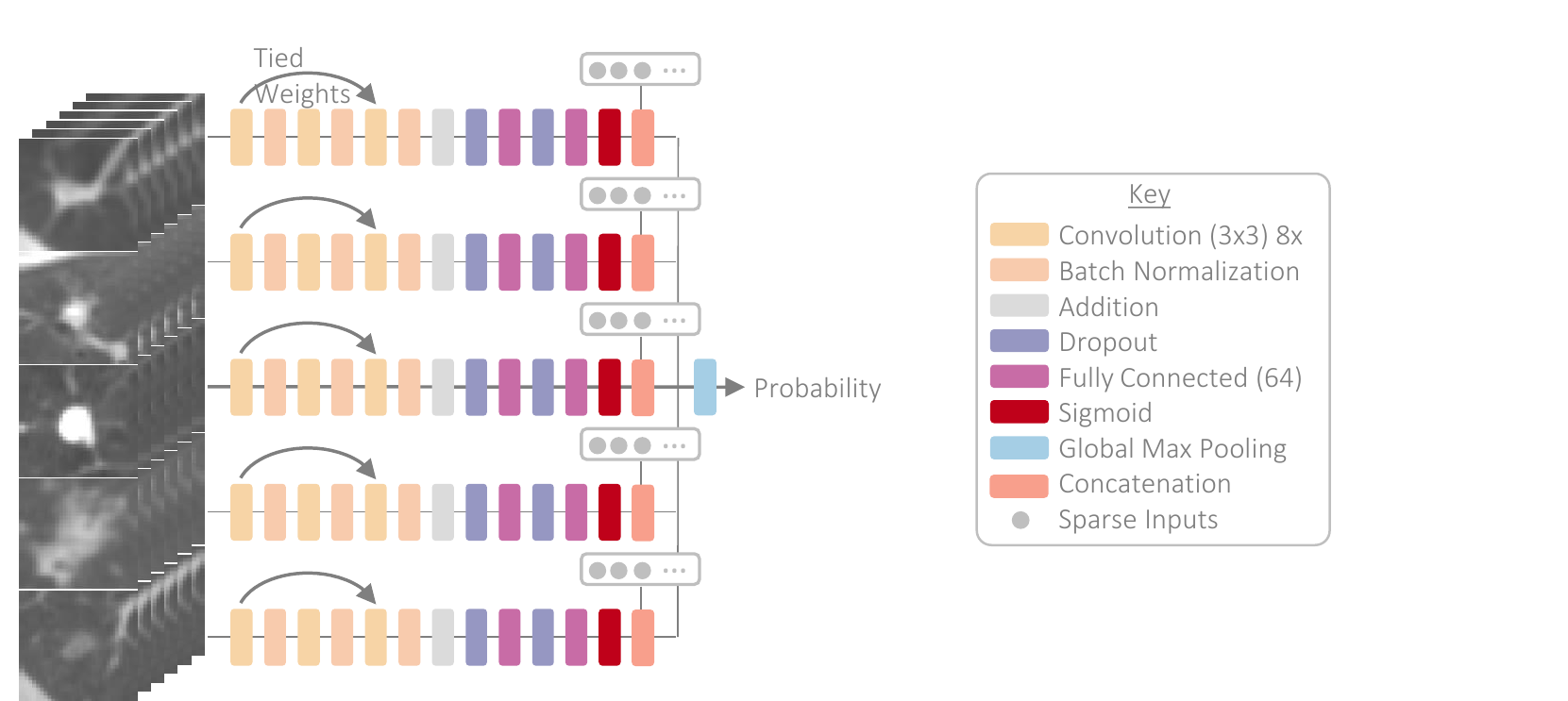}
    \end{subfigure}
\caption{A: The pipeline of our algorithm. Initially a nodule detector is used to identify the nodules contained in a CTLS scan. Subsequently, the ten largest nodules are provided as input to a deep learning algorithm that assesses the cancer risk of the scan. B: The network architecture that is applied to the cube around the detected nodules.}\label{fig:Our_algorithm_pipeline}
\end{figure*}

\emph{Model architecture.} We propose a two-stage machine learning framework for cancer risk assessment. In the first stage, we employ a nodule detector to identify the nodules that are contained in a CTLS scan while in the second stage we use the ten largest nodules identified by the nodule detector as input to a deep and wide neural network that assesses their cancer risk. The decision to use the ten largest nodules was based on the optimal performance obtained from experiments with different numbers of nodules used as input. The details of the two stages are given in the remainder of this section. The pipeline of the algorithm is shown in Figure~\ref{fig:Our_algorithm_pipeline}.

% add Figure 1 here

We evaluate this framework with two nodule detectors, one based on Hierarchical Machine Learning by Bergtholdt \emph{et al.}~\cite{Bergtholdt_etal_Hamburg_detector_2016} and another based on deep neural network semantic segmentation by Liao \emph{et al.}~\cite{grt123team_kaggle2017}. Both are based on LUNA16 dataset~\cite{LUNA16} of confirmed nodule cases, while the later contains additional annotated cases of very big nodules (or already developed tumors) from the Kaggle dataset. More details about the nodule detectors are given in the above mentioned manuscripts~\cite{grt123team_kaggle2017,Bergtholdt_etal_Hamburg_detector_2016}.

The nodule detector provides us with the nodule locations in all three dimensions: $x$, $y$ and $z$ as well as additional information such as the nodule size (e.g., radius in mm), and the confidence of the suggestion - given by the nodule detector. We refer to these parameters as nodule metadata.

Based on the output from the previous stage, we can extract from the CTLS scan localized cubes of 32x32x32mm$^3$ around a nodule (and since we employ isotropic resampling to 1mm$^3$ each voxel corresponds to 1mm$^3$). This gives us sufficient context for the experiments as we find that smaller or larger cubes do not improve and can even degrade performance. Additionally, during training, a random crop of 28x28x28mm$^3$ out of the extracted 32x32x32mm$^3$ cube is taken to ensure that the network does not see the same images in each batch iteration thus reducing overfitting. Finally, from the 3D 28x28x28mm$^3$ cube we extract three different 2D projections, as channels, namely coronal, sagittal, and transverse, thus ending up with 3 times 28x28 input per nodule for the neural network (Figure~\ref{fig:Our_algorithm_pipeline}).

\begin{table*}[h!tb]
\centering
\caption{Architecture of the deep and wide neural network.}
\label{table:NeuralNet}
\begin{tabular}{lrlcc}
        \multicolumn{1}{c}{Layer} & \multicolumn{1}{c}{Properties} & Previous layer(s)  \\
\toprule\toprule
1. Image input & (10x3x28x28)          & - Img: 10 nod, 3proj 28x28            \\
2. Conv Layer + BN & (3x3, 8x), stride 1                     & 1.                                                          \\
3. Conv Layer + BN      & (3x3, 8x), stride 1                   & 2.                                         \\
4. Conv Layer + BN      & (3x3, 8x), stride 1                    & 3.                                           \\
5. Conv Layer + BN      & (3x3, 8x), stride 1                    & 1.                                           \\            
6. Addition/Merge + BN      & -                    & 4., 5.                                           \\          
7. Dropout + BN      &                     & 6.                                           \\  
8. Dense + BN      & (64)                   & 7.                                           \\          
9. Dropout + BN      &                     & 8.                                           \\  
10. Dense + BN      & (64)                   & 9.                                           \\          
11. Numeric input & (10x1)          & -  Radius   \\
%12. Numeric input & (10x1)          & -  Sphericity   \\
12. Numeric input & (10x1)          & -  $x$, $y$, $z$ nodule coordinates   \\
13. Numeric input & (10x1)          & -  confidence score   \\
14. Addition/Merge      & -                    & 10., 11., 12., 13.                                           \\
15. Dense + sigmoid      & (1)                   & 14.                                           \\  
16. GlobalMaxPool      & (10)                   & 15.                                           \\  
\toprule
\end{tabular}
\end{table*}

% add Table 2 here

Moreover, for each nodule we use additional features, such as nodule radius, and confidence score (confidence level of a detected nodule as provided by the algorithm used for nodule detection) as numeric inputs added in the penultimate level in the architecture. The nodule descriptors are obtained automatically by the nodule detector without any human intervention. In the experiments that use the SVM-based nodule detector~\cite{Bergtholdt_etal_Hamburg_detector_2016} we employ one additional feature, nodule sphericity, that is not provided by the Deep Learning based nodule detector~\cite{grt123team_kaggle2017}. Different volumes have different number of nodules. In the experiments, we used the 10 largest nodules, when there are at least 10 nodules in the volume, otherwise all the nodules are used and the remaining “spots” are masked. We use a ResNet-like~\cite{He_etal_2016} deep and wide neural network for evaluating the cancer risk associated with each CTLS scan. (Deep refers to the number of layers, while wide refers to the number of inputs.) The input consists of the image part as described in the previous paragraph and the additional nodule features (e.g., radius etc.) of the nodule properties added at the penultimate layer. The network architecture is visualized in Figure~\ref{fig:Our_algorithm_pipeline}. More details of the exact layer configuration of the neural network are given in Table~\ref{table:NeuralNet}. We used 3x3 kernels for convolutional neural network blocks with 8 channels with stride 1, intertwined with batch normalization and additional connections for realizing the ResNet-blocks (see inputs 5. and 6. in Table~\ref{table:NeuralNet}), augmented with dropout for better generalization and followed by fully connected layers (with 64 units) and sigmoid activation functions. Finally, we concatenate the last fully connected layer with the nodule metadata, making the deep neural network also wide. At the end, we perform a global max pooling aggregating over the maximum of ten branches representing the different nodules, which estimates the final cancer risk probability. We have employed the aforementioned architecture with both nodule detectors, with the only difference being the dropout rate. More precisely, when using the SVM-based nodule detector we set the dropout rate to very high values (0.7-0.9) that were necessary to obtain optimal performances, while for the Deep Learning based nodule detector~\cite{grt123team_kaggle2017} the dropout rate was set to a much smaller value (0.25).

\hspace{-1.2em}\emph{Training of our model and performance evaluation.} Our model relies on information about verified cancer diagnosis at the volume/scan level. This implies that our CT volumes were annotated with label 1 in cases where the patient was diagnosed with lung cancer and 0 otherwise. In this sense our data can be categorized as multi-instance weakly labeled, since our labels (cancer diagnosis) are provided for the group of nodules that are contained within a scan and not for each nodule individually.

\begin{figure*}[h!b]
    \centering
    \begin{subfigure}[b]{0.4\textwidth}
        \centering
        \includegraphics[width=\textwidth]{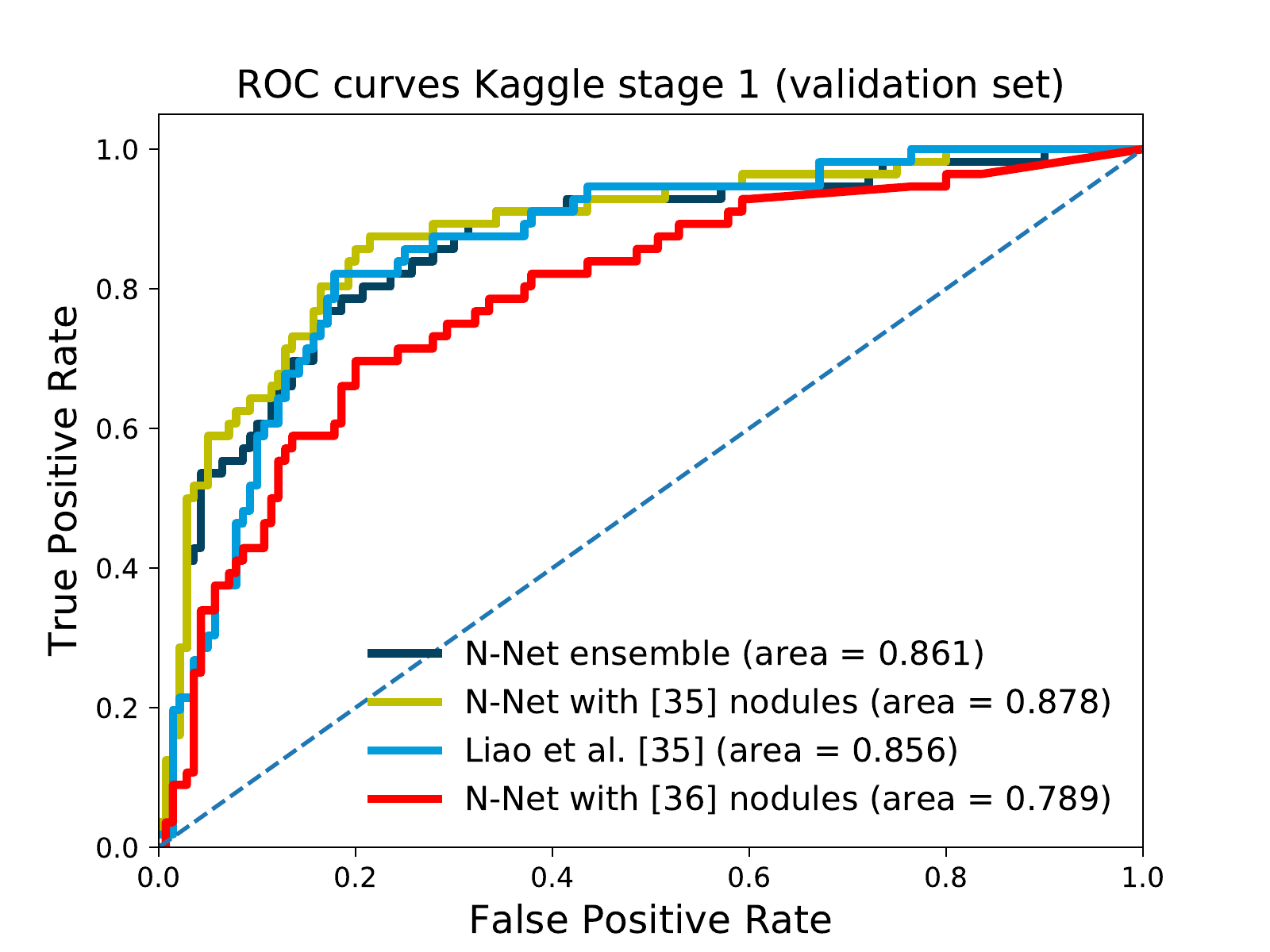}
        \caption*{}
    \end{subfigure}
		\begin{subfigure}[b]{0.4\textwidth}
        \centering
        \includegraphics[width=\textwidth]{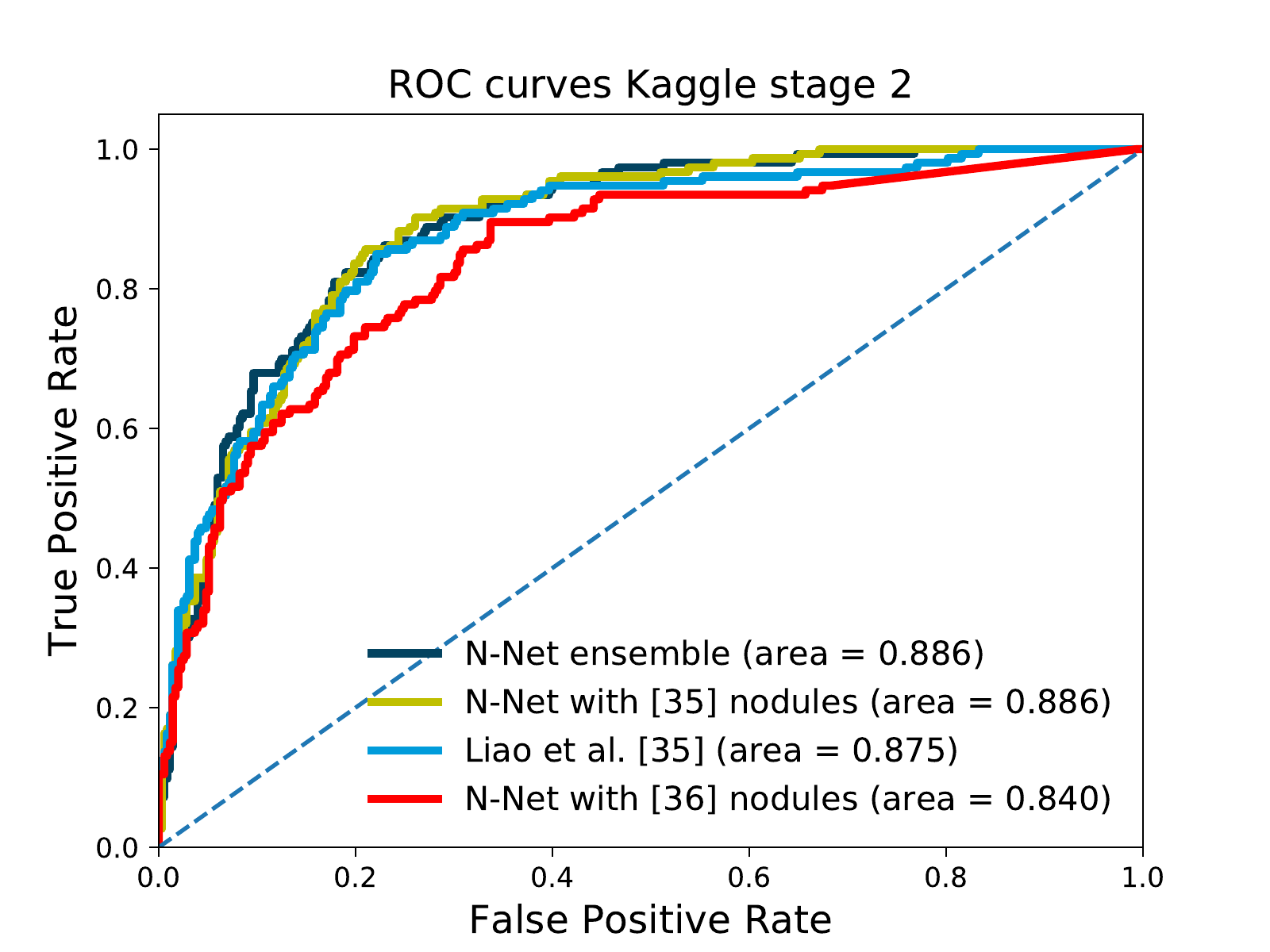}
        \caption*{}
    \end{subfigure}

    \begin{subfigure}[b]{0.4\textwidth}
        \centering
        \includegraphics[width=\textwidth]{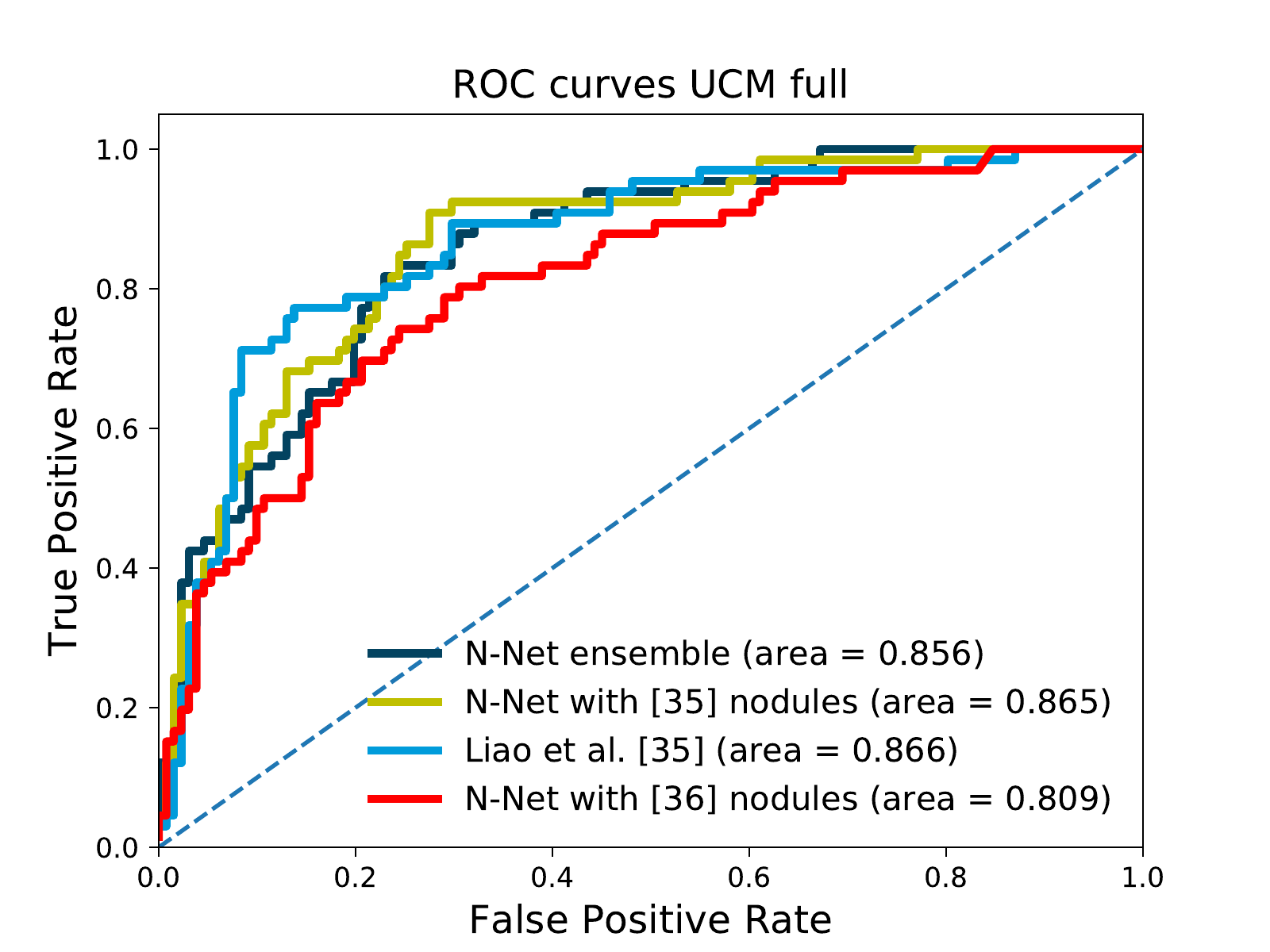}
        \caption*{}
    \end{subfigure}
		\begin{subfigure}[b]{0.4\textwidth}
        \centering
        \includegraphics[width=\textwidth]{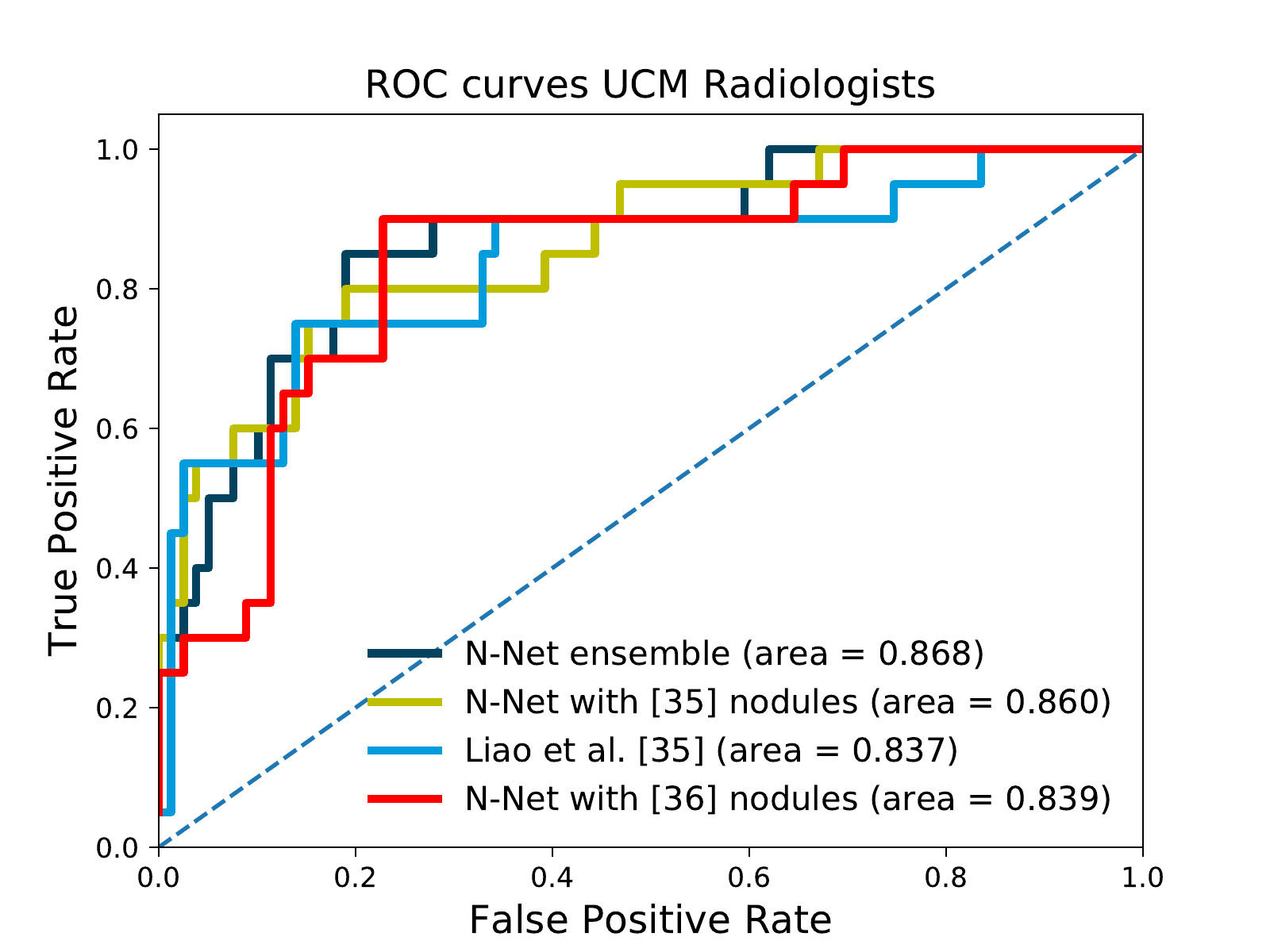}
        \caption*{}
    \end{subfigure}

    \begin{subfigure}[b]{0.4\textwidth}
        \centering
        \includegraphics[width=\textwidth]{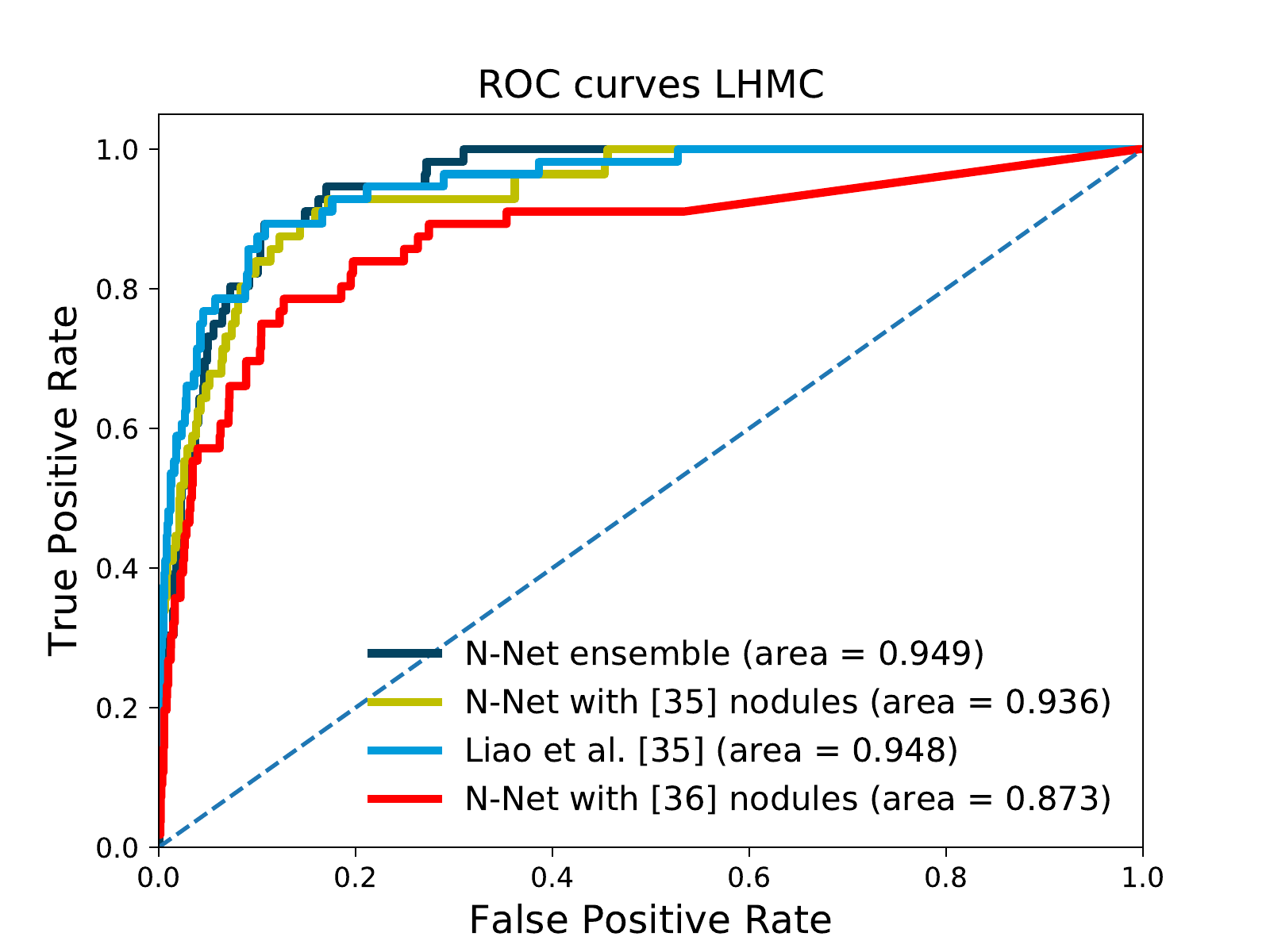}
        \caption*{}
    \end{subfigure}
		\begin{subfigure}[b]{0.4\textwidth}
        \centering
        \includegraphics[width=\textwidth]{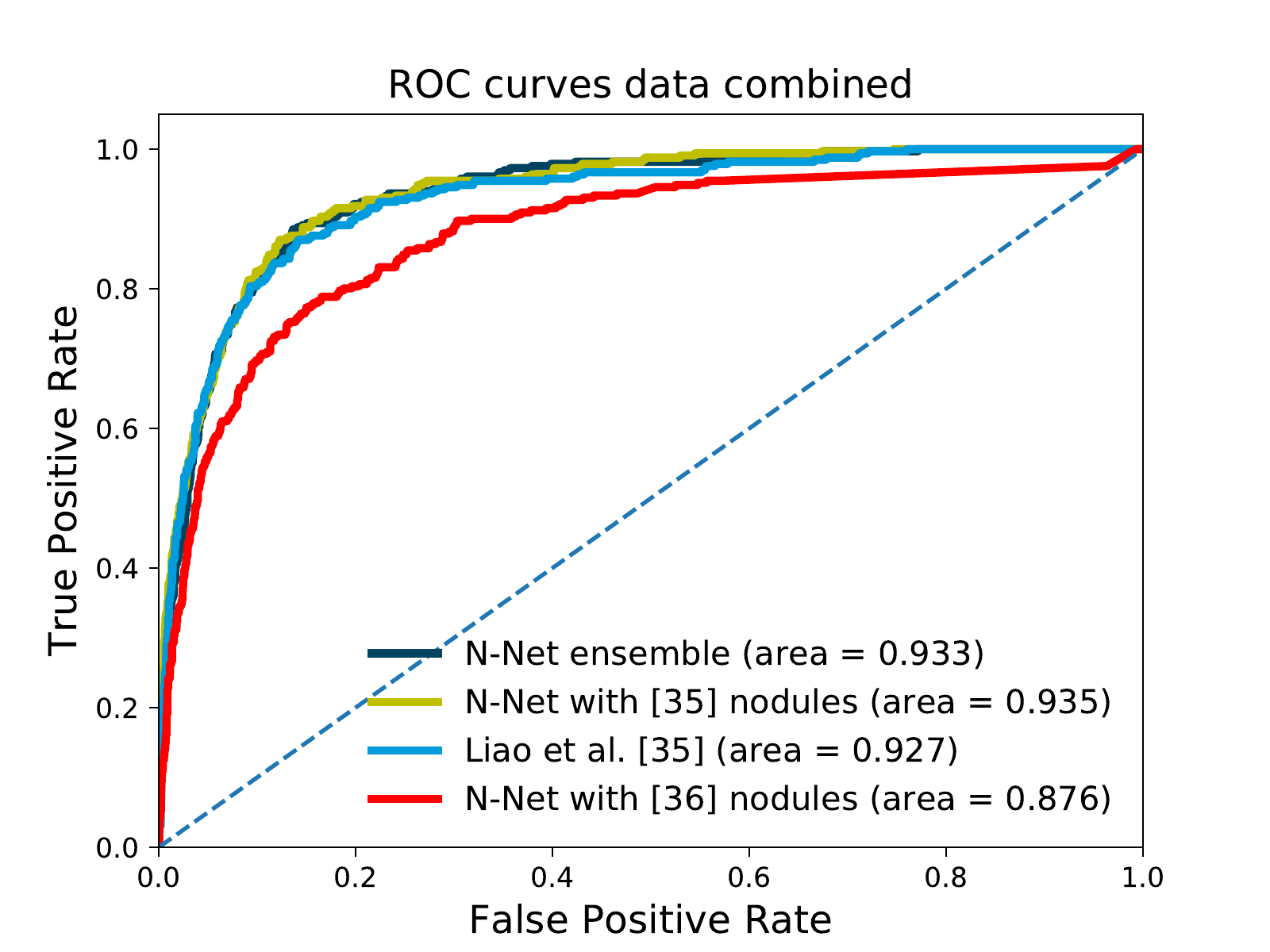}
        \caption*{}
    \end{subfigure}

\caption{Performance of our model is consistent across different datasets. We use the acronym Nodule-Net (N-Net) to refer to our neural network architecture described in Table~\ref{table:NeuralNet}. For the N-Net model that employs the nodules of~\cite{grt123team_kaggle2017} we have trained the model using 5-fold cross validation on NLST, excluding the patients of the UCM dataset. In all figures during inference, N-Net with~\cite{grt123team_kaggle2017} nodules is applied as the average of these 5-folds. For the N-Net model that employs the nodules of~\cite{Bergtholdt_etal_Hamburg_detector_2016} we have used 90\% of the NLST dataset and set to 0.9 the dropout rate to train the model that is reported at the LHMC and Kaggle subfigures, while for the UCM full \& radiologists subfigures we have employed the full NLST dataset for training excluding the relevant test-set patients. For the UCM full and radiologists’ subset the N-Net dropout rate was set to 0.8 and 0.7 respectively. On the combined dataset AUCs for N-Net and Liao \emph{et al.} are 93.5\% and 92.7\%, respectively with a p-value $P=.0446$. To compare any two classifiers, we use one-sided permutation test~\cite{everitt2010cambridge} for computing the p-value.}\label{fig:ROC_curves_comparison_datasets}
\end{figure*}
% Figure 2

This information was available in all datasets reported in Table~\ref{table:Data_used}. Using these labels at the volume level, we trained our neural network with the binary cross-entropy loss function. In the empirical results, we always evaluate the performance of our model with respect to verified cancer diagnosis at the volume level. The model that employs the DL-based nodule detector of~\cite{grt123team_kaggle2017}, was trained using 5-fold cross-validation on the NLST data excluding all patients used in the UCM dataset and during inference the average prediction of the 5-folds was used. For the SVM-nodule model (that employs nodule detector of~\cite{Bergtholdt_etal_Hamburg_detector_2016}) we have employed again the NLST data for training and we have used three model versions, one trained using 90\% of the NLST training data and 0.9 dropout, one using the full NLST data excluding the full UCM patient list (197 patients) and 0.8 dropout and the full NLST data excluding the UCM subset rated by Radiologists (99 patients) and 0.7 dropout. This was done because we observed that the SVM-nodule model had reduced performance on test data (like the Kaggle dataset) when using smaller NLST subsets for training. Thus, when testing the model performance we employed the larger version possible from NSLT, i.e. we used the NLST dataset without the UCM patients only when testing for that data subset. This behavior was not observed when we employed the DL-nodule model and we were able to use a single model excluding all UCM patients from NLST without observing reduced performance in our test sets. Moreover, we should note that the convergence of our models when using high dropout rate (even 0.9) was consistent, possibly due to the fact that we used the nodule metadata at the penultimate layer of the neural network architecture. We used Adam optimizer~\cite{Adam_KingmaBa} with a learning rate of $1e-3$.

We explored the individual contribution of the architectures attributes with various ablation experiments. Performance comparison of the different neural network architectures using the SVM-based nodule detector are given in an extended previous version of the manuscript\footnote{\href{https://arxiv.org/abs/1804.01901v1}{https://arxiv.org/abs/1804.01901v1}}. Namely, we tried using small to moderate dropout, using less or more global nodule features (e.g., goodness, brightness, Hounsfield units (HU), $x$, $y$ and $z$ nodule dimensions), using only a single (largest) nodule, taking larger or smaller part around a nodule, using different architectures such as VGGs~\cite{Simonyan14c} and DenseNets~\cite{Huang_et_al_2017}. The results suggest that there is no benefit in these architectures and the proposed one (in Table~\ref{table:NeuralNet} and Figure~\ref{fig:Our_algorithm_pipeline}) performs better than the alternative architectures or hyper-parameters. Similar hyper-parameter exploration was conducted for the DL-based nodule detector, resulting in the same neural network architecture with the only difference being the optimal dropout rate, which was lower (0.25 for the~\cite{grt123team_kaggle2017} nodule detector vs. 0.7-0.9 for the~\cite{Bergtholdt_etal_Hamburg_detector_2016} nodule detector).

\subsection*{PanCan Risk Model}
To empirically validate our framework, we employ a model developed at the Vancouver General Hospital for nodule malignancy estimation~\cite{McWilliams_Vancouver_2013}. This method provides information using a single scan, and does not use information potentially available from multiple scans of the patient (that could be used, for example, to identify nodule growth). The model employs a formula, which calculates the malignancy score based on 9 numerical or boolean input parameters, including three patient features: age of a patient, gender of a patient, lung cancer family history (true or false); one clinical or image-based feature: presence of emphysema (true or false); one patient specific image-based feature: number of nodules in the CTLS scan; and four nodule specific image-based features: size of a nodule (diameter) - which is longest in-slice axis, type of the nodule (nonsolid, part-solid, or solid), location of the nodule in the upper lobe (true or false), and nodule spiculation (true or false).

% add equation 1 here
\begin{equation}
\text{nodule malignancy score} = \frac{1}{1+e^{-\sum \limits_{i=1}^{9} \text{weight}_i \cdot \text{input}_i}}
\end{equation}
To compare our model that produces a single risk score for each CTLS scan to the \emph{PanCan risk model} that computes a risk score on a per-nodule basis, we set the CTLS scan malignancy score to be derived by the maximum malignancy score of all nodules. In our experiments this provides the best performance results for the \emph{PanCan risk model} (rather than taking the mean, minimum scores etc. of a nodule per study).

\subsection*{Radiologists Predictions}
To compare our results to radiologist performance, an observer study was conducted at UCM using 99/197 CTLS scans for which radiologists have provided a continuous numeric estimate of the cancer probability in addition to the Lung-RADSTM score. This subset consists of 20 malignant and 79 benign cases. Each selected case had to have at least one nodule within the range of 6-25mm. Besides nodule size distribution matching, the selection covered nodule types of all categories except for calcified nodules. Three senior and three junior radiologists from the thoracic imaging department participated in the study. A graphical user interface was designed for the study to capture and demonstrate relevant information to the user such as the three orthogonal views (axial, sagittal, and coronal) of the imaging focused on the slices containing the nodule as well as demographic information such as sex, age, smoking history, and family history of smoking. The user was able to measure the nodule size using the measurement tool provided. After taking all information into account, the radiologist was asked to provide the assessment of the risk for developing lung cancer in terms of a percentage number.

\begin{figure*}[h!tb]
    \centering
    \begin{subfigure}[b]{0.4\textwidth}
        \centering
        \caption{}\label{fig:Lahey_ourmodel_vs_Vancouver_ROC}
        \includegraphics[width=\textwidth]{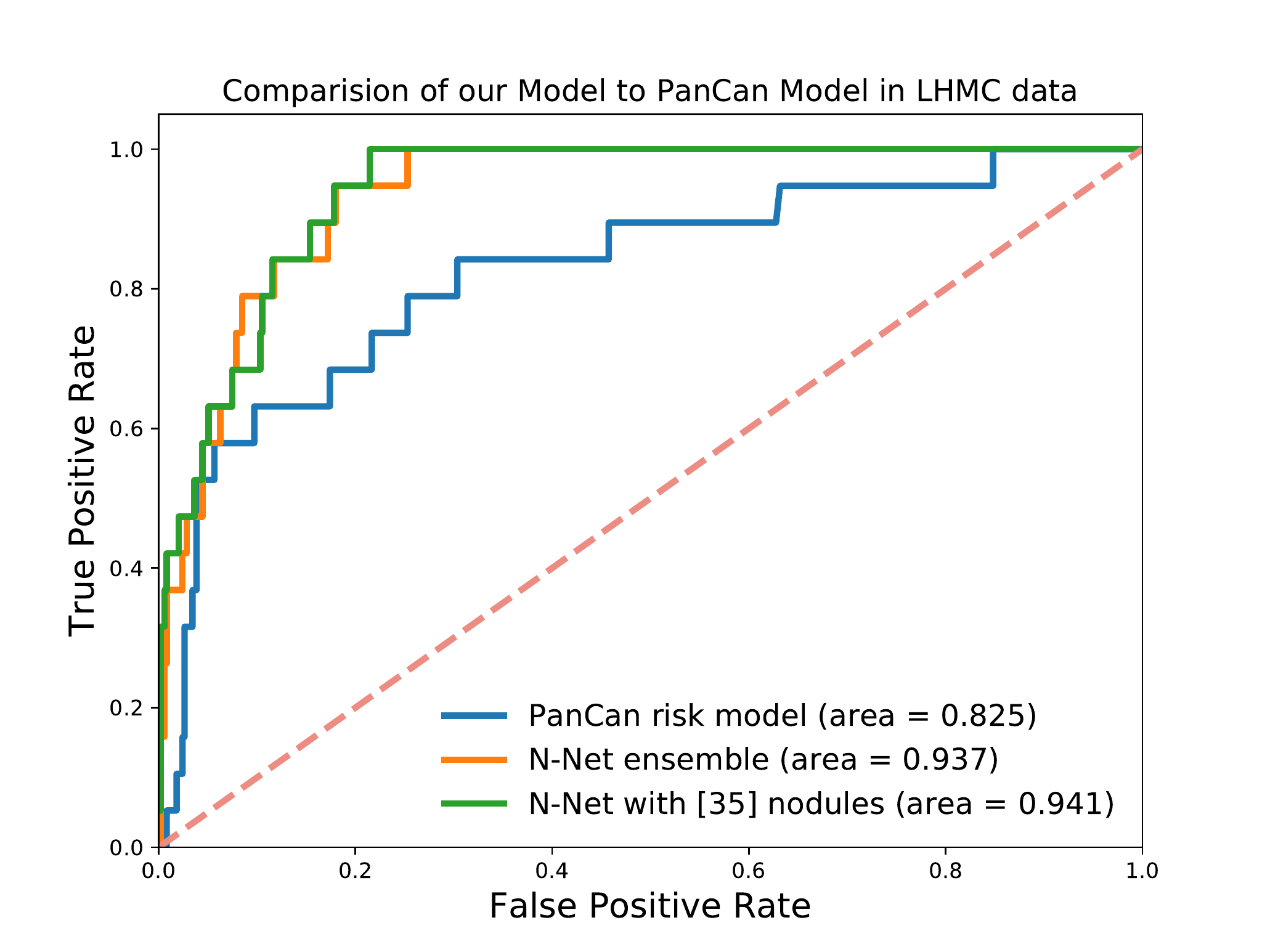}
    \end{subfigure}
		\begin{subfigure}[b]{0.4\textwidth}
        \centering
\caption{}\label{fig:UCM_ourmodel_vs_Vancouver_ROC}
        \includegraphics[width=\textwidth]{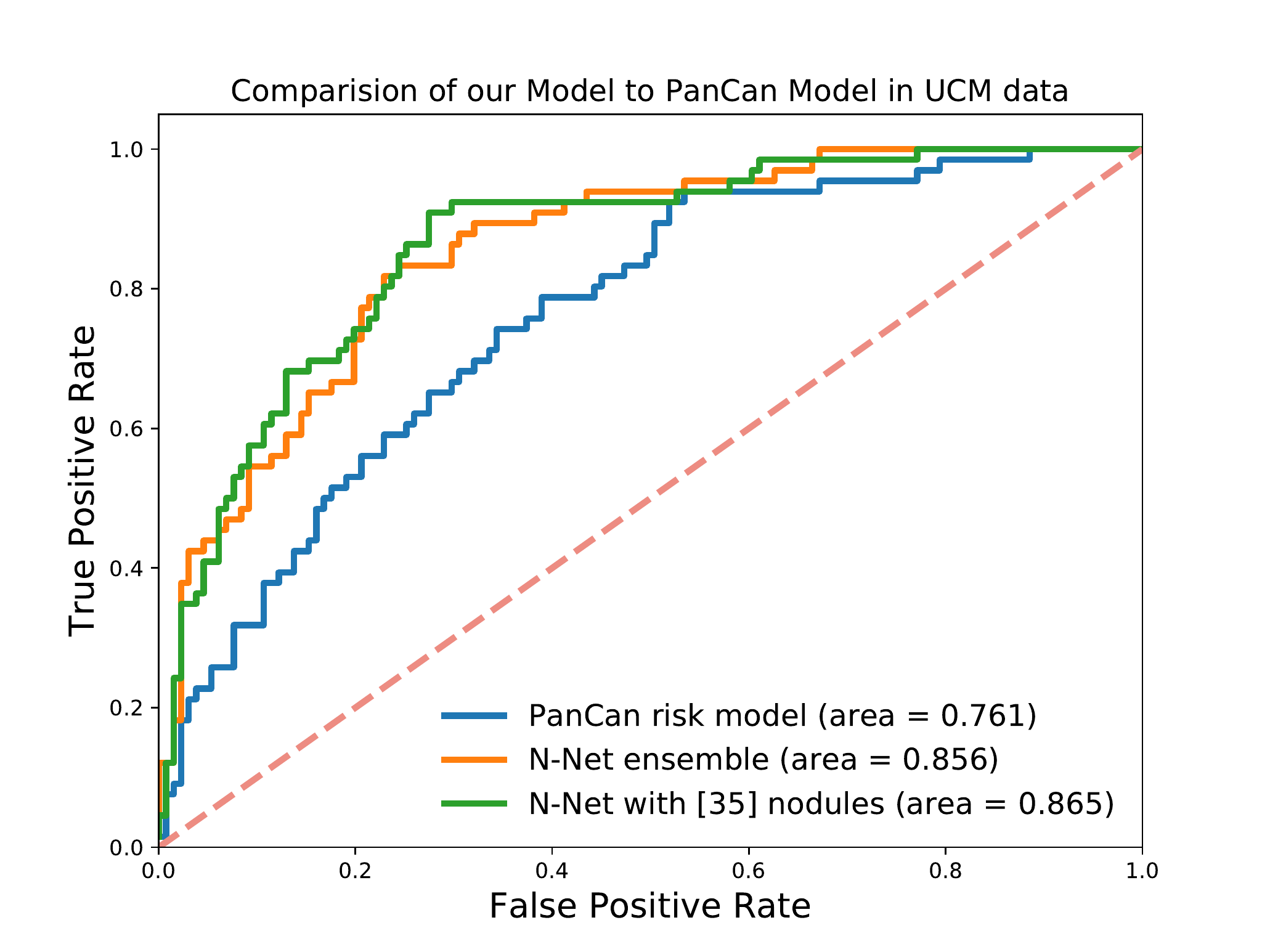}       
    \end{subfigure}

    \begin{subfigure}[b]{0.4\textwidth}
        \centering
\caption{}\label{fig:UCM_ourmodel_vs_radiologists_ROC}
        \includegraphics[width=\textwidth]{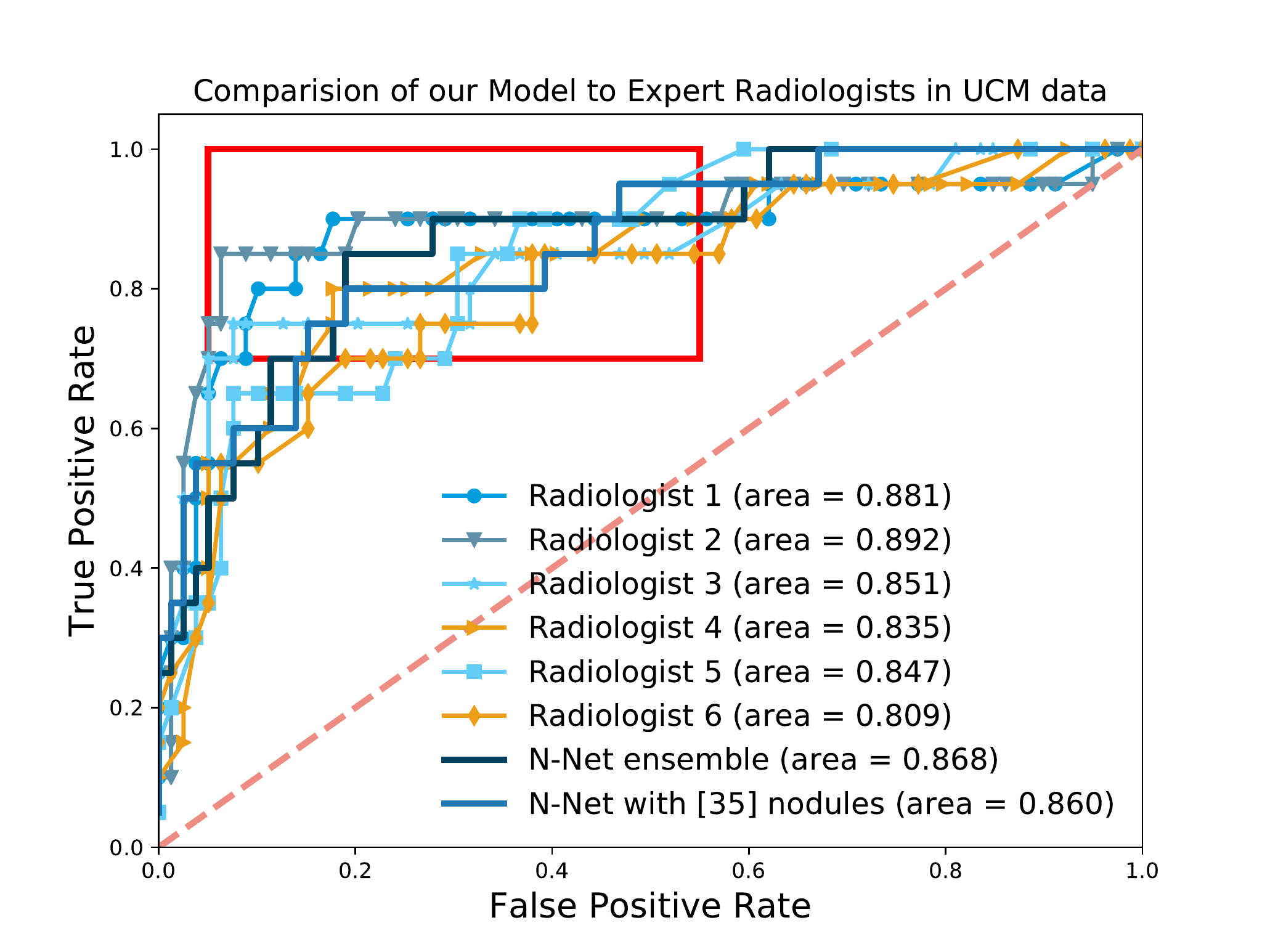}
    \end{subfigure}

\begin{subfigure}[b]{0.23\textwidth}
        \centering
\caption{}\label{fig:Lahey_performance_sensitivity_leveled_specificity}
        \includegraphics[width=\textwidth]{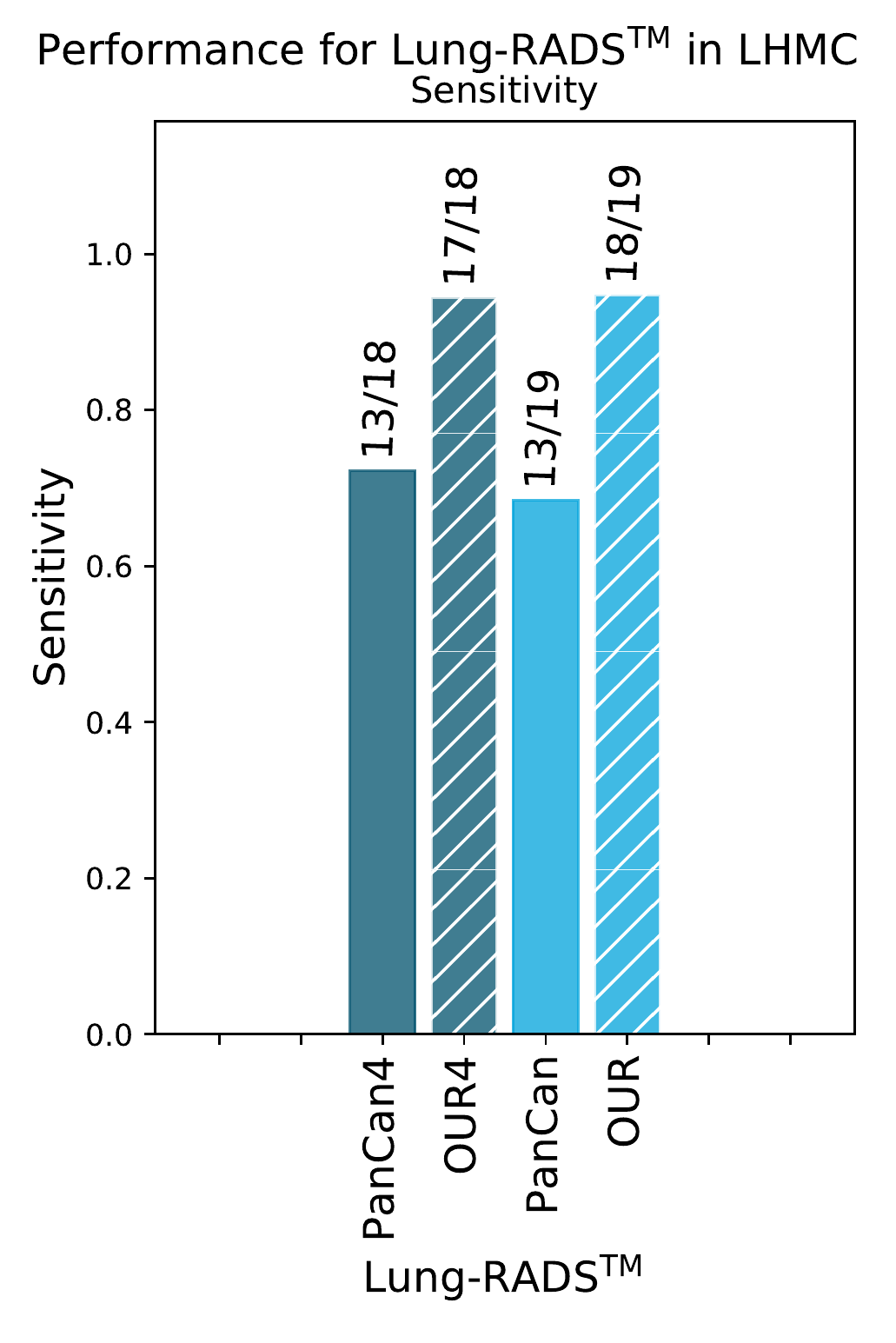}
    \end{subfigure}
 \begin{subfigure}[b]{0.23\textwidth}
        \centering
 \caption{}\label{fig:Lahey_performance_specificity_leveled_sensitivity}
        \includegraphics[width=\textwidth]{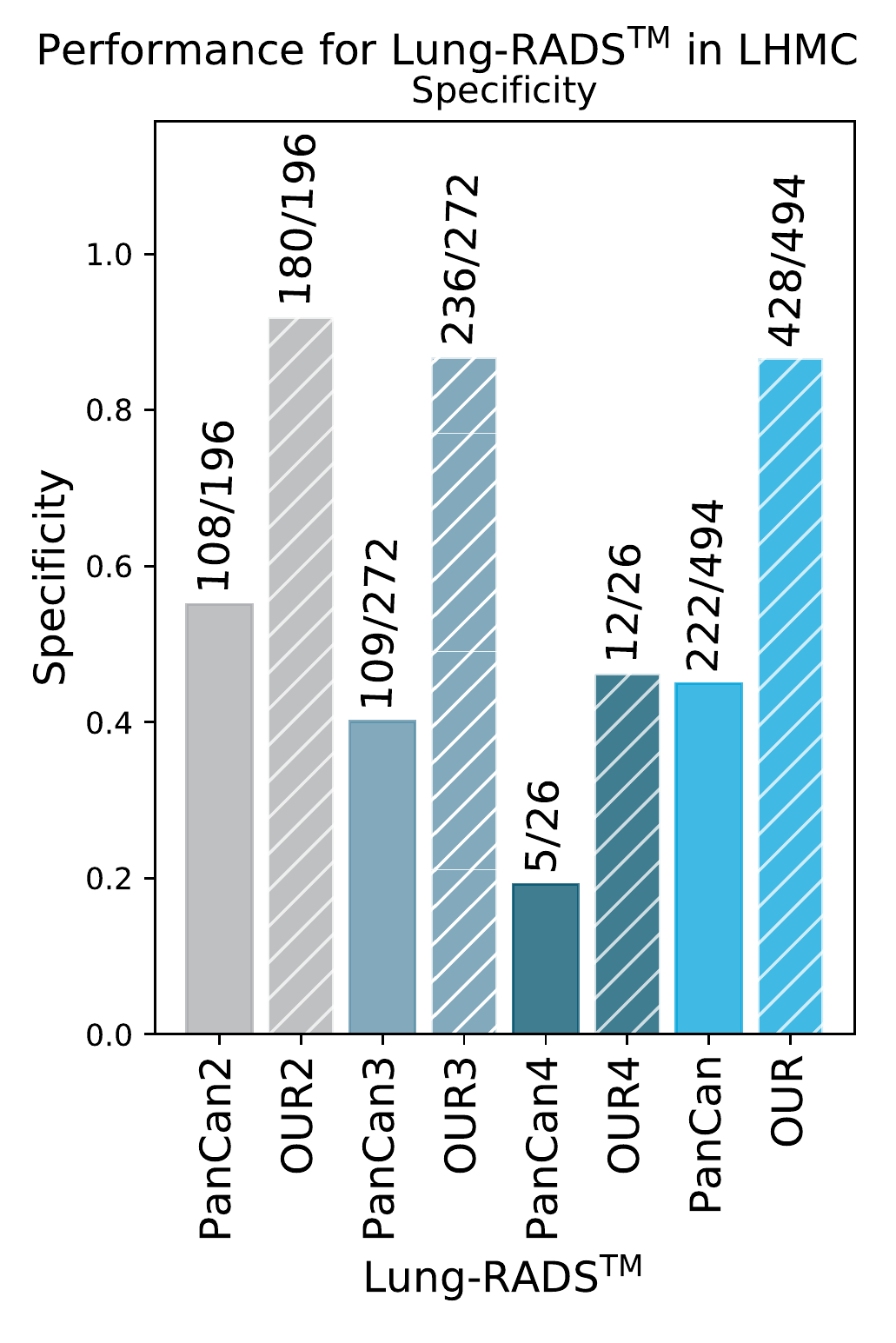}     
    \end{subfigure}

\caption{Lung cancer risk assessment performance of our DNN model compared to \emph{PanCan risk model}~\cite{McWilliams_Vancouver_2013} and radiologists for UCM and LHMC data. A: ROC curve showing the performance of our model and the \emph{PanCan risk model} for LHMC data. B: ROC curve showing the performance of our model and the \emph{PanCan risk model} for all 197 studies in UCM data (with 64 verified cancer cases). C: ROC curve showing the performance of our model compared to radiologists' assessments for 99 studies that have available annotations in UCM data. D: Lung-RADS\textsuperscript{TM} grouped sensitivity for LHMC data when the specificity is set to 80\%;  E: Lung-RADS\textsuperscript{TM} grouped specificity for LHMC data when the sensitivity is set to 84\%. ``OUR2,3,4" and ``PanCan2,3,4" labels refer to the performance achieved for Lung-RADS\textsuperscript{TM} = 2,3,4 classified cases, while ``PanCan" and ``OUR" refer to the N-Net model as described in Table~\ref{table:NeuralNet}, trained using the~\cite{grt123team_kaggle2017} nodules on the NLST dataset excluding the UCM patients.}\label{fig:comparison_vancouver_radiologists}
\end{figure*}
% Figure 3

\section*{Results}

\subsection*{Performance of the algorithm and comparison with the state-of-the-art}

The performance of our framework was stable across the different datasets used and can achieved an AUC (Area Under the Curve) score between 86\%-94\% as shown in Figure~\ref{fig:ROC_curves_comparison_datasets}. It is worth re-iterating that our model has been trained using only data from one dataset (NLST), but generalizes well across all different datasets that we used in the experiments. Our evaluation is more extensive than the majority of related works that commonly use smaller and less diverse datasets. The model is trained on NLST dataset and it does not require additional re-training and is just evaluated on the remaining data sets. The results also show that the performance is better than the winner of the Kaggle Data Science Bowl challenge on Lung Cancer Screening, is better than the widely used \emph{PanCan risk model} and is comparable to a panel of six radiologists.

\subsection*{The influence of the choice of the nodule detection}

In Figure~\ref{fig:ROC_curves_comparison_datasets}, it can be observed that although the same neural network architecture (referred to as N-Net in the figures) was used with both nodule detectors of~\cite{grt123team_kaggle2017} and~\cite{Bergtholdt_etal_Hamburg_detector_2016}, there is a substantial performance difference. This can be attributed to the fact that the SVM-based nodule detector~\cite{Bergtholdt_etal_Hamburg_detector_2016} missed a malignant nodule in some cases, especially where the nodule was already a large developed tumor (thus outside the range of nodules covered by LIDC data, where~\cite{Bergtholdt_etal_Hamburg_detector_2016} was trained on). These cases, although small in number, significantly affected the ROC curves since these cancer cases were associated with a very low probability.

\subsection*{Comparison with radiologists performance}
Figure~\ref{fig:UCM_ourmodel_vs_radiologists_ROC} shows the ROC curves of our model as compared to the ROC curves obtained by the single-scan risk assessments of the 6 radiologists on the subset of 99 volumes (different patients) of the UCM data out of which 20 correspond to verified cancer cases. Our algorithm shows a comparable and often better performance than the performance of the radiologists. The predictions of the two radiologists that have slightly better AUC scores, namely Radiologists 1 and 2 in Figure 3C, have p-values~\cite{everitt2010cambridge}, when compared to our methods, way higher from what is normally considered statistically significant ($P=.386$ and $P=.299$, respectively).

We highlight with a red box the area of the ROC curve where the true positive rate is at a high level, which is an important factor when performing LCS (i.e. no cancer cases are missed). It should be noted that our work is one of the few studies~\cite{vanRiel_et_al2017} in the literature where such a comparison to radiologists is performed.

\subsection*{Comparison with the PanCan Risk Model}
The results, presented in Figures~\ref{fig:Lahey_ourmodel_vs_Vancouver_ROC} and \ref{fig:UCM_ourmodel_vs_Vancouver_ROC} show that our proposed model significantly outperforms the \emph{PanCan risk model}~\cite{McWilliams_Vancouver_2013} by approximately 11\% AUC for both UCM and LHMC datasets with p-values~\cite{everitt2010cambridge} $P=.0146$ and $P=.0537$, respectively. Further, we compare our algorithm with the \emph{PanCan risk model} for various Lung-RADSTM categories in the LHMC data. We performed the evaluation by comparing the sensitivity for different fixed specificity performance levels and vice versa (i.e. comparing specificity for fixed levels of sensitivity for both algorithms. These evaluations per different Lung-RADSTM categories show that our algorithm performs better than the \emph{PanCan risk model} in terms of sensitivity and specificity (in Figure~\ref{fig:Lahey_performance_sensitivity_leveled_specificity} and \ref{fig:Lahey_performance_specificity_leveled_sensitivity}).

\section*{Discussion}

As we have discussed in detail in the introduction section, there are several research papers that have proposed solutions for the problems of nodule detection and nodule malignancy assessment. However, the evaluation of these models is done usually in datasets of much smaller scale that what is used in this paper. For example~\cite{Anirudh_et_al_2016} uses a dataset of 300 CT scans to train and evaluate their model while in this study we employ 20x more scans from a combination of public and private data sources. This allows us to have more confidence about the robustness and generalization capacity of our framework. Moreover, we evaluated the performance of the DSB Kaggle competition winners~\cite{grt123team_kaggle2017} in all our datasets, thus allowing us to make a thorough comparison of our model performance against state-of-the-art approaches. Interestingly~\cite{grt123team_kaggle2017} had strong performances in most datasets thus illustrating the robustness of modern deep learning approaches and also the usefulness of data challenges and competitions to advance scientific research. 

From the methodological point of view, our work contributes to a better understanding of the type of information that is needed to train highly performant Deep Learning models for cancer malignancy estimation. The two main paradigms are (i) models that employ solely nodule location information and patient outcomes (diagnosed-cancer or no-cancer) as information to train the model, (ii) models that employ additional information about nodule characteristics (such as level of spiculation, lobulation, etc.). We employ only the nodule locations and patient diagnostic outcomes to train our model. To the extent of our knowledge the only other work that relies on the same information is~\cite{grt123team_kaggle2017}, i.e. the Kaggle DSB winner’s model that scored better against approaches that employed additional nodule characteristics. The results presented in this paper illustrate that nodule locations and patient diagnostic outcomes suffices to build high-performant deep learning models for lung cancer malignancy estimation.

Another interesting methodological question is related to whether nodule detection and malignancy estimation should be trained in a common end-to-end model or they can be two separate tasks. The model proposed in~\cite{grt123team_kaggle2017} is trained in an end-to-end manner, while in our work we have used a two-stage model where the training of the second stage is separate from the nodule detection. Interestingly, our results demonstrate that a two-stage approach can be very successful and both nodule detectors of~\cite{grt123team_kaggle2017} and~\cite{Bergtholdt_etal_Hamburg_detector_2016} performed very well on datasets that they have not seen before (like NLST and LHMC). Our results can lead to interesting follow-ups, since patient diagnosis results can be retrieved from patient records and do not have the level of ambiguity that is sometimes associated with radiologist reports. Moreover, the availability of a substantial number of nodule detectors as a result of the LUNA16 challenge means that a clinical research center that performs lung cancer screening has the potential to build and validate their own lung cancer screening model based on the principles outlined in this work.

\section*{Conclusion}

Lung cancer malignancy risk assessment is an important research topic that has recently attracted a lot of attention due to the fact that there are nearly 10,000,000 people in the US alone that fit the high-risk criteria for CTLS. This illustrates the need to develop tools to help radiologists evaluate the CTLS scans and protect the patients without lung cancer from the risks associated with unnecessary care escalation.

In this paper, we propose a two-stage framework for cancer risk assessment that is shown to have (i) robust performance across three low-dose CT dataset, (ii) improved performance compared to SOTA models and (iii) comparable performance to a panel of six radiologists. As a focus for further work, one can consider the differences in model performance across different image quality settings such as reconstruction filters (soft-tissue, sharp, etc.). One can potentially improve performance by limiting the neural networks' training and subsequently the prediction on a unique set of reconstruction filters or consider domain adaptation methods to optimize performance across different image quality data.

%\bibliography{./bibliography}

%\section*{Acknowledgements}% (not compulsory)
%
%We would like to thank ...

%Acknowledgements should be brief, and should not include thanks to anonymous referees and editors, or effusive comments. Grant or contribution numbers may be acknowledged.
%\vspace{-1em}
\section*{Author contributions}
%S.T., D.M., C.L.S., B.G.G., and B.S.V. designed and implemented the deep neural networks part of the algorithm. R.W. and T.K. designed and implemented the nodule detector part of the algorithm. S.T., D.M., C.L.S., B.G.G., and B.V. designed, implemented and conducted the experiments and analyzed the results. C.L.S., A.T., S.M.R., C.W., B.J.M. and H.M. participated in the data gathering or provided part of the data. S.T. and D.M. wrote the manuscript. H.P. and D.M. supervised the research. All authors reviewed and approved the manuscript.

Mr. Trajanovski and Mr. Mavroeidis had full access to the data and can take responsibility for the integrity of the data and the accuracy of the data analysis. Mr. Trajanovski and Mr. Mavroeidis affirm that the manuscript is an honest, accurate, and transparent account of the study being reported; that no important aspects of the study have been omitted; and that any discrepancies from the study as planned (and, if relevant, registered) have been explained.

\hspace{-1.4em}\emph{Concept and design:} Trajanovski, Mavroeidis, Leon Swisher, Gebrekidan Gebre

\hspace{-1.4em}\emph{Acquisition, analysis, or interpretation of data:} Trajanovski, Mavroeidis, Leon Swisher, Gebrekidan Gebre, Veeling, Wiemker, Klinder, Tahmasebi, Regis, Wald,  McKee, Flacke, MacMahon, Pien.

\hspace{-1.4em}\emph{Statistical analysis:} Trajanovski, and Mavroeidis.

\hspace{-1.4em}\emph{Drafting of the manuscript:} Trajanovski, Mavroeidis, and Leon Swisher

\hspace{-1.4em}\emph{Critical revision of the manuscript for important intellectual content:} Trajanovski, Mavroeidis, Leon Swisher, Gebrekidan Gebre, Veeling, Wiemker, Klinder, Tahmasebi, Regis, Wald,  McKee, Flacke, MacMahon, and Pien
Supervision: Mavroeidis, and Pien.

%\vspace{-0.7em}
\section*{Additional information}

\textbf{Competing financial interests}. The authors declare no competing financial interests.

\clearpage

\onecolumn

\end{document}